\documentclass[11pt]{article}
\usepackage{booktabs}

\usepackage[final]{acl}

\usepackage{times}
\usepackage{latexsym}
\usepackage{amsmath}
\usepackage{todonotes}
\usepackage{xurl}
\usepackage{hyperref}
\usepackage{enumerate,enumitem}
\usepackage{lipsum}
\usepackage[T1]{fontenc}

\usepackage[utf8]{inputenc}

\usepackage{microtype}
\usepackage{algorithm}
\usepackage{algorithmicx}
\usepackage{algpseudocode}
\usepackage{pifont}
\usepackage{inconsolata}

\usepackage{graphicx}
\usepackage{xcolor}
\usepackage{gradient-text}
\usepackage{xspace}
 \usepackage{multirow} 
\usepackage{listings}
\usepackage[table]{xcolor}
\definecolor{lightblue}{RGB}{235,242,249}
\definecolor{lightred}{RGB}{250,230,235}
\definecolor{jpblue}{RGB}{33, 99, 154}
\definecolor{jpred}{RGB}{188, 0, 45}

\newcommand{\model}{\gradientRGB{FinReporting}{0,114,178}{213,94,0}\xspace}

\title{\model: An Agentic Workflow for Localized Reporting of Cross-Jurisdiction Financial Disclosures}

\author{
 \textbf{Fan Zhang\textsuperscript{1,2}\thanks{zhang-fan@g.ecc.u-tokyo.ac.jp}} \ 
 \textbf{Mingzi Song\textsuperscript{3}} \
 \textbf{Rania Elbadry\textsuperscript{1}} \
 \textbf{Yankai Chen\textsuperscript{1,4}}\thanks{corresponding author: yankaichen@acm.org} \
 \textbf{Shaobo Wang\textsuperscript{2}} \\
 \textbf{Yixi Zhou\textsuperscript{1}} \
 \textbf{Xunwen Zheng\textsuperscript{5}} \
 \textbf{Yueru He\textsuperscript{6}} \
 \textbf{Yuyang Dai\textsuperscript{7}} \
 \textbf{Georgi Georgiev\textsuperscript{8}} \\
 \textbf{Ayesha Gull\textsuperscript{9}} \
 \textbf{Muhammad Usman Safder\textsuperscript{9}} \
 \textbf{Fan Wu\textsuperscript{1}} \
 \textbf{Liyuan Meng\textsuperscript{1}} \
 \textbf{Fengxian Ji\textsuperscript{1}} \\
 \textbf{Junning Zhao\textsuperscript{2}} \
 \textbf{Xueqing Peng\textsuperscript{10}}\footnotemark[1] \
 \textbf{Jimin Huang\textsuperscript{10}} \
 \textbf{Yu Chen\textsuperscript{2}} \
 \textbf{Xue (Steve) Liu\textsuperscript{1,4}} \\
 \textbf{Preslav Nakov\textsuperscript{1}} \
 \textbf{Zhuohan Xie\textsuperscript{1}} \\
 \textsuperscript{1}MBZUAI \
 \textsuperscript{2}The University of Tokyo \
 \textsuperscript{3}Meiji Gakuin University \
 \textsuperscript{4}McGill University \\
 \textsuperscript{5}Kyoto University \
 \textsuperscript{6}Columbia University  \
 \textsuperscript{7}University of California, Berkeley \\
 \textsuperscript{8}Sofia University “St. Kliment Ohridski” \
 \textsuperscript{9}Namal University \
 \textsuperscript{10}The Fin AI
}

\begin{document}
\maketitle

\begin{abstract}
Financial reporting systems increasingly leverage Large Language Models (LLMs) to extract and summarize corporate disclosures. However, most existing approaches assume a single-market setting and overlook structural differences across jurisdictions. Variations in accounting taxonomies, tagging infrastructures (e.g., XBRL vs.\ PDF), and aggregation conventions introduce substantial challenges for semantic alignment and reliable verification. Here, we aim to bridge this gap. 
We present \model, an agentic workflow for localized cross-jurisdiction financial reporting. The system constructs a unified canonical ontology spanning the income statement, balance sheet, and cash flow statement, and decomposes reporting into auditable stages, including filing acquisition, extraction, canonical mapping, and anomaly logging. Rather than treating LLMs as free-form generators, \model employs them as constrained verifiers operating under explicit decision rules with evidence grounding.
Evaluated on annual filings from the USA, Japan, and China, \model improves consistency and reliability under heterogeneous reporting regimes. We further release an interactive demo that enables cross-market inspection and supports structured export of localized financial statements.
Our demo is available at \url{https://huggingface.co/spaces/BoomQ/FinReporting-Demo}.
A video describing our system is available at \url{https://www.youtube.com/watch?v=f65jdEL31Kk}.

\end{abstract}

\section{Introduction}

Financial statements are indispensable for investors to assess a firm's financial condition and performance. 
With recent breakthroughs in Large Language Models (LLMs), a growing line of work has begun to automate customized financial reporting.

Given a user's analytical focus, systems parse long filings together with relevant tables, and extract or summarize the financial facts that matter most to that user~\cite{huang2024extracting,aguda2024large,wang2025can,shu2025lava,wang2026fintaggingbenchmarkingllmsextracting}. 
This reduces the burden of reading complete reports, and makes financial disclosures more accessible for decision-making.

\begin{figure}[t]
  \centering
  \includegraphics[width=\columnwidth]{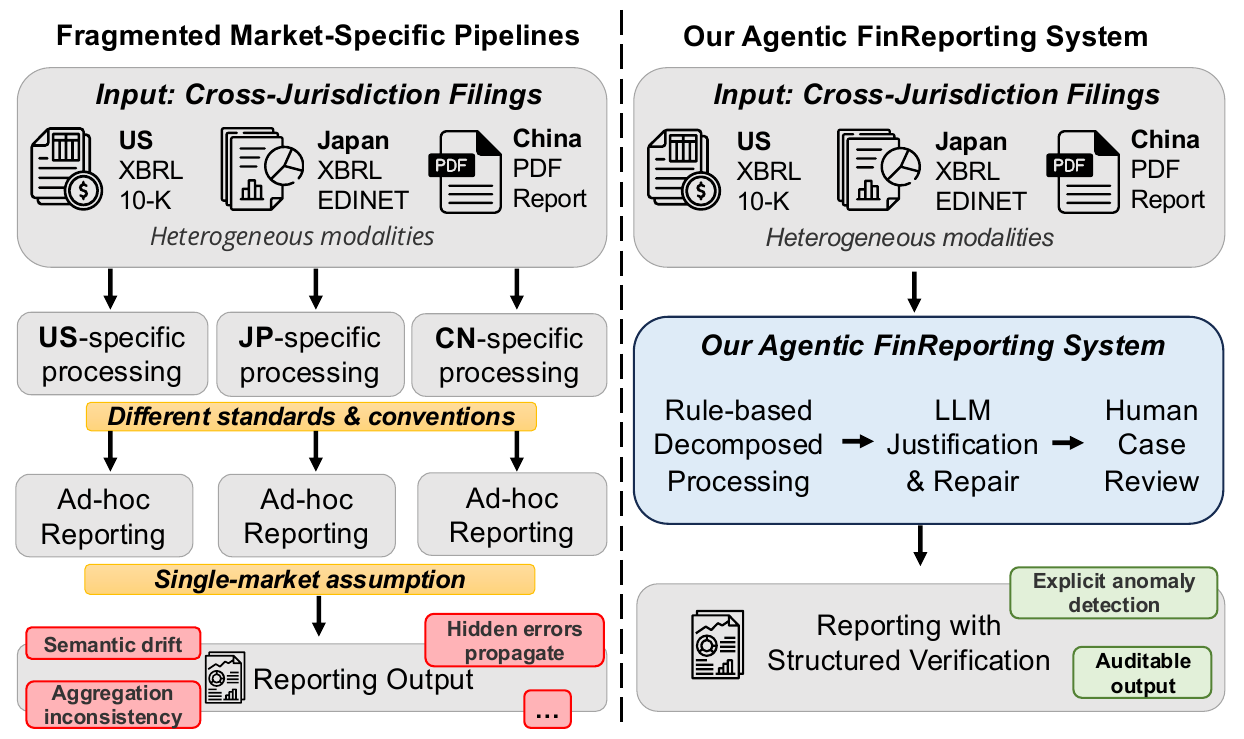}
  \caption{
  Financial reporting is traditionally handled via separate pipelines operating under the \textit{single-market} assumption, leading to implicit issues. Our \model~system specifically implements the localized reporting of cross-jurisdiction financial disclosures. 
  }
  \label{fig:overview}
\end{figure}

However, existing systems predominantly operate under a \textit{single-market assumption}: the user queries filings within the same jurisdiction where they are already familiar with the accounting standards, disclosure conventions, and reporting requirements, as shown in the left-hand side of Figure~\ref{fig:overview}. 
For global investors, this assumption breaks. 
When an investor attempts to understand a foreign firm's financial statements, two practical frictions arise. 
On the one hand, national financial infrastructures, e.g., auditing regimes, taxonomy standards, and consolidation conventions, can vary substantially across jurisdictions. 
Thus, investors who are not deeply familiar with the target market often struggle to interpret raw filings correctly. 

On the other hand, financial data vendors, such as Bloomberg\footnote{\url{https://www.bloomberg.com}} and Wind,\footnote{\url{https://www.wind.com.cn}} rely heavily on manual extraction and expert validation to curate structured financial data. Although this ensures high accuracy, it is labor-intensive, costly, and difficult to scale, leading to high subscription costs and limited accessibility, particularly for small and mid-sized investors.
Therefore, these realities motivate the need for an automatic framework that can \textit{restructure cross-jurisdictional financial data into a localized representation aligned with the investor's home-market logic}.


However, cross-jurisdictional reporting localization extends beyond translation or format conversion. Divergent accounting taxonomies and aggregation conventions mean that superficially similar line items may represent different concepts, while equivalent concepts may appear under different labels, leading to semantic drift and inconsistent roll-ups. Because such misalignments can remain hidden and propagate into downstream quantitative models, reliable localization must explicitly support structured verification and repair rather than treating filings as visual document parsing.

To fill this gap, we propose \model, an agentic workflow for unified, localized reporting of cross-jurisdiction financial disclosures.
\model~constructs a canonical financial ontology based on universal reporting standards, e.g.,~Income Statement (IS), Balance Sheet (BS), and Cash Flow (CF), to align semantically equivalent items across markets.
In this work, we implement this ontology across United States (US), Japanese (JP), and Chinese (CN) markets.
Operationally, \model~decomposes end-to-end localization into a sequence of auditable steps before \textit{Outputting}, including \textit{Filing Acquisition}, \textit{Statement Identification}, \textit{Extraction}, \textit{Canonical Mapping}.
Moreover, \model~includes a structured verification mechanism between steps.
Specifically, at each step, it combines rule-based constraints with LLM-based reasoning to validate, repair, and justify intermediate outputs, ensuring that resulting localized statements are both logically coherent and semantically faithful to source filings.
Therefore, this enables \model~to go beyond separate information processing and surface anomalies in information aggregation for financial reporting, as shown in the right-hand side of Figure~\ref{fig:overview}.

\begin{figure*}[t]
  \centering
\includegraphics[width=\textwidth]{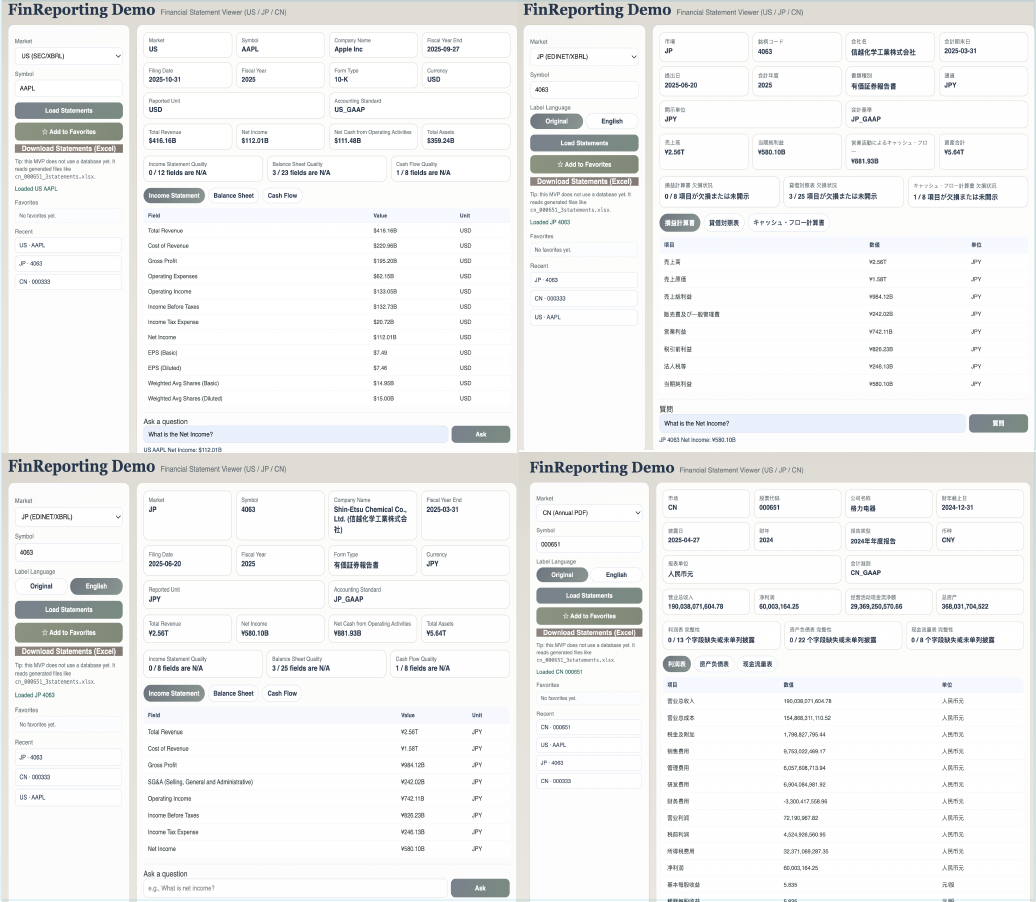}
  \caption{
  The \model demo interface: the system supports cross-jurisdiction statement browsing and reporting (US/JP/CN), with specific interested fields selected.
  }
  \label{fig:cross-column-fig}
\end{figure*}

As illustrated by the demo interface in Figure~\ref{fig:cross-column-fig}, \model enables users to load market-specific filings by symbol, view extracted IS/BS/CF statements, and export localized financial reports.
It provides an auditable infrastructure for downstream applications, including financial question answering, regulatory analysis, and cross-market benchmarking. 
By operationalizing structured verification and canonical alignment, it enables transparent and consistent cross-jurisdiction financial reporting.
Our contributions can be summarized as follows:
\begin{itemize}[leftmargin=*]
    
    \item We present \model{}, a system for localized reporting of cross-jurisdiction financial disclosures across heterogeneous reporting standards and formats, enabling unified canonical alignment and consistent cross-market interpretation.

    \item We implement an auditable agentic workflow that integrates rule-based extraction, ontology-guided canonicalization, and constrained LLM-based verification and repair, producing structured outputs with explicit audit trails, anomaly flags, and quality signals (see Figure~\ref{fig:pipeline}).
    

    \item We develop an interactive interface that allows users to explore localized financial statements, inspect verification evidence and audit trails, and export structured reports for transparent analysis and downstream applications.

\end{itemize}

\section{Related Work}

\subsection{Financial NLP}
Financial NLP has long studied textual signals in corporate disclosures, such as sentiment, tone, and linguistic cues for market prediction and risk assessment~\citep{loughran2011liability,kogan2009predicting,qian2025agentstradelivemultimarket,fincards}.
More recently, the field has expanded toward numerical reasoning and question answering over financial reports, with benchmarks such as FinQA, TAT-QA, and conversational extensions~\citep{chen2021finqa,zhu2021tatqa,chen2022convfinqa,finchain,qian2025fino1transferabilityreasoningenhancedllms, ji2025finestate}, as well as retrieval-centric long-form settings, e.g., FinTextQA~\citep{chen2024fintextqa}.
In parallel, finance-oriented foundation models and evaluation suites, e.g., BloombergGPT, FinGPT, FinBen, MultiFinBen, have been proposed to strengthen domain adaptation and capability measurement~\citep{wu2023bloomberggpt,yang2023fingpt,xie2024finben,peng2025multifinbenbenchmarkinglargelanguage}.


\subsection{Financial Disclosure Reporting}
Modern financial disclosure systems increasingly adopt XBRL, i.e., standardized digital tags for financial data, to enable machine-readable reporting. 
For instance, the U.S. SEC provides large-scale structured datasets derived from these tags, allowing direct access to reported facts~\citep{sec_fsn_datasets,sec_fsdatasets}.
Building on such infrastructures, recent research has proposed pipelines for extracting and structuring tagged financial facts, including LLM-oriented end-to-end benchmarks for financial information extraction and structuring~\citep{wang2026fintaggingbenchmarkingllmsextracting} and instruction-tuned models for taxonomy-scale extreme classification~\citep{khatuya2024parameter}.
Recent LLM-based interfaces for querying XBRL-tagged disclosures, e.g.,~XBRL-centered analysis agents, further improve usability~\citep{xbrlagent2024}.

However, disclosure practices and reporting standards vary substantially across markets and regulatory environments worldwide.
Although many jurisdictions have introduced digital reporting mechanisms, their taxonomies and reporting conventions are not directly interchangeable~\citep{ifrs_taxonomy_2024,esma_esef_taxonomy_2024_documentation, ji2026servimageimagegenerationediting}.
As a result, most existing pipelines are based on the \emph{single-market} setting, as they assume a fixed taxonomy and reporting logic, limiting scalability and generalization across jurisdictions.
Unlike existing XBRL-centric or single-market extraction pipelines that assume fixed taxonomies and homogeneous infrastructures, \model explicitly models cross-jurisdiction heterogeneity and embeds structured verification, thus enabling robust, scalable, and auditable localization across heterogeneous filing regimes.

\section{\model~System}

\begin{figure}[t]
  \centering
  \includegraphics[width=\columnwidth]{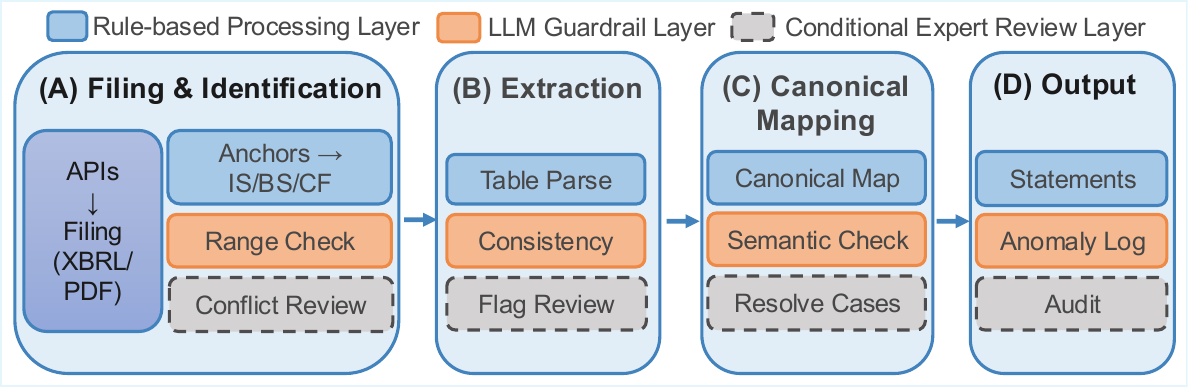}
  \caption{
  Overview of the \model processes. 
  }
  \label{fig:pipeline}
\end{figure}

\model~is an agentic workflow that partitions financial reporting into auditable steps: \textit{Filing Acquisition}, \textit{Statement Identification}, \textit{Extraction}, and \textit{Canonical Mapping}, followed by the final \textit{Outputting} step.
Between steps, \model~inserts structured verification mechanisms, i.e., LLM justification and human-expert review, to ensure intermediate output quality.
Thus, as shown in Figure~\ref{fig:pipeline}, the system adopts a three-layer design to sequentially execute these steps:
\ding{182} a deterministic rule-based processing layer that produces reproducible initial results,
\ding{183} an LLM verification/repair layer with strict guardrails, and
\ding{184} a lightweight human review and evaluation layer for high-impact cases.

\subsection{Rule-based Processing Layer}
\paragraph{Filing Acquisition and Statement Identification.}~\model~first acquires the target annual filing and identifies the relevant statement sections with the given query.
The acquisition strategy is jurisdiction-aware: for XBRL-native (US/JP) markets, \model~directly loads tagged facts from the filing; for PDF-centric (CN) markets, it locates the corresponding annual report and detects the page ranges of the core statements such as IS/BS/CF.
This step identifies a statement-level context package (including document pointers, period metadata, and statement boundaries) that serves as the shared input to downstream extraction.

\paragraph{Extraction.}~\model~then proceeds with rule-based parsers to generate stable, explainable, and reproducible candidate values without relying on LLM generation.
We distinguish two extraction tracks because the input regimes are fundamentally different.
In XBRL-native jurisdictions (US/JP), filings already provide standardized, machine-readable tags; extraction mainly reduces to \emph{selecting the correct facts} under reporting context, e.g., consolidated vs separate, period length, and instant/duration matching, and exporting the tagged items into our statement schema.

In PDF-centric jurisdictions (CN), disclosures are not consistently tagged and table layouts vary across issuers. Therefore, extraction requires \emph{document-level decomposition}, table parsing, robust column selection with fallbacks, and per-field status labeling.
In this work, we additionally attach a per-field status label to each extracted item to explicitly surface uncertainty, such as \texttt{OK}, \texttt{MISSING}, \texttt{PARSE\_ERROR}, and \texttt{NOT\_APPLICABLE}.

\paragraph{Canonical Mapping via a Global Ontology.}
In order to localize statements into an investor-home representation, \model~maps the extracted items into a unified canonical schema guided by a global financial ontology spanning IS/BS/CF.
The ontology defines a set of core concepts and their cross-market correspondences, enabling semantically aligned structuring across heterogeneous filings.
Canonical mapping produces localized statement tables under the same concept inventory, which makes cross-jurisdiction comparison and downstream applications (e.g., QA and benchmarking) consistent.

\paragraph{Outputting.}

Finally, \model\ outputs (\emph{i})~localized financial statements under the unified canonical schema, and (\emph{ii})~an anomaly log and audit trail that record the irregularities and decisions encountered in previous steps.

\subsection{LLM Guardrail Layer}
To prevent \emph{undetected mistakes}, i.e., plausible-looking outputs that are incorrect but not flagged, \model~treats the LLM as a \emph{bounded verifier} rather than a free-form extractor.
At each stage, \model~combines pre-defined constraints with LLM-based reasoning to (\emph{i})~validate intermediate outputs, (\emph{ii})~propose repairs when evidence is sufficient, and (\emph{iii})~justify decisions with traceable support.
The verifier operates under a restricted decision space: \texttt{KEEP} (retain the rule value), \texttt{REPAIR} (override with an evidence-backed value), or \texttt{NEED\_REVIEW} (defer to human).
A repair is applied only when all guardrails pass: the field is repairable (typically \texttt{MISSING} or \texttt{PARSE\_ERROR}), the evidence is explicitly grounded in the filing context, and the proposed value is consistent with the cited evidence; 
otherwise, the system falls back to \texttt{NEED\_REVIEW}.
All decisions are logged with evidence and failure reasons, ensuring that the final localized statements remain logically coherent and semantically faithful to the source filings.

\subsection{Conditional Expert Review Layer}
\model\ supports targeted human review for cases flagged as \texttt{NEED\_REVIEW} or large discrepancies.
Reviewers inspect audit trails and evidence to resolve conflicts.
For demonstration, we provide structured templates and ablation comparisons to quantify how verification and repair affect extraction completeness and localization consistency.


\begin{table}[t]
\centering
\small
\setlength{\tabcolsep}{3pt}
\begin{tabular}{ll|cc}
\toprule
\textbf{Jurisdiction} & \textbf{Metric} & $\mathrm{LLM}_{\mathrm{Reporting}}$ & \model \\
\midrule
\multirow{3}{*}{US-Jurisdiction} 
    & FR  & 94.44 & 95.56 \\
    & CR  & 5.56  & 15.56 \\
    & ACC & 89.38    & 90.23    \\
\midrule
\multirow{3}{*}{JP-Jurisdiction} 
    & FR  & 84.44 & 84.44 \\
    & CR  & 15.56 & 15.56 \\
    & ACC & 88.36    & 88.36    \\
\midrule
\multirow{3}{*}{CN-Jurisdiction} 
    & FR  & 63.33 & 63.33 \\
    & CR  & 26.67 & 40.56 \\
    & ACC & 78.15    & 82.11    \\
\bottomrule
\end{tabular}
\caption{Performance across jurisdictions. In this experiment, we use \texttt{GPT-4o} in our LLM guardrail layer.}
\label{tab:jurisdiction_results}
\end{table}

\section{Experiments}

\subsection{Setup}
\paragraph{Evaluation Protocol.}
We evaluate \model over core financial statement fields from annual filings. 
For each company, the system produces reporting for a fixed inventory of eighteen core items spanning IS/BS/CF, with an auditable trace and a review flag when evidence is insufficient.

\paragraph{Evaluation Data.}
We evaluate \model under a unified annual-filing protocol across the US, JP, and CN, focusing on non-financial firms (excluding banks, insurers, and securities firms due to different reporting schemas), consolidated statements, and canonical IS/BS/CF mapping. Data are collected from public regulatory disclosures: US filings from SEC EDGAR (10-K XBRL),\footnote{\url{https://www.sec.gov/search-filings}} JP filings from EDINET annual securities reports (XBRL),\footnote{\url{https://disclosure2.edinet-fsa.go.jp/}} and CN annual reports from publicly disclosed PDFs.\footnote{\url{http://www.cninfo.com.cn/}}
For each market, we construct a 20-company baseline split for rule development and gold annotation, and a 10-company challenge split as a held-out evaluation set, yielding 30 firms per market (90 total). Gold annotations are manually validated by financial experts to ensure correct canonical mappings and numerical consistency.

Cross-market metrics are reported on a shared subset of 18 canonical targets (IS: 5, BS: 7, CF: 6), while market-specific items are retained for UI display and qualitative analysis.


\paragraph{Evaluation Metrics.}
Let $N$ be the total number of target fields. We report
(\emph{i})~\emph{Filled Rate (FR)}, the fraction of fields with non-null outputs,
(\emph{ii})~\emph{Conflict Rate (CR)}, the fraction of fields triggering human review due to disagreement between the deterministic result and the LLM verifier (or insufficient evidence),
and (\emph{iii})~\emph{Accuracy (Acc)}, computed against manual labels over the reviewed fields.

\subsection{Empirical Analysis}

\paragraph{Comparison to Na\"{i}ve LLM Reporting Pipeline.}

As shown in Table~\ref{tab:jurisdiction_results}, the performance is strongest for US examples, slightly lower for Japanese, and substantially weaker for Chinese for both methods. This pattern is largely driven by differences in data structure and standardization. US disclosures are highly machine-readable and schema-consistent (e.g.,~standardized tags and stable semantics), making the task closer to structured extraction and mapping; Japan follows a similar paradigm, but shows greater variability in tagging and reporting, increasing cross-firm alignment complexity. In contrast, China often requires extracting and reconstructing information from PDF reports, where layout variability, table fragmentation, and inconsistent line items introduce noise and amplify error propagation, resulting in the lowest reliability.

\begin{table}[t]
\centering
\small
\setlength{\tabcolsep}{3pt}
\begin{tabular}{l|rrr|r}
\toprule
\textbf{LLM Backbones} & \textbf{FR} & \textbf{CR} & \textbf{ACC} & \textbf{Cost ($\$$)} \\
\midrule
\texttt{GPT-5.2} & 95.56 & 8.89 & 90.23 & 36.96 \\
\texttt{GPT-5 mini} & 95.56 & 15.00 & 90.23 & 17.77 \\
\texttt{GPT-4o} & 95.56 & 15.56 & 90.00 & 34.04 \\ 
\texttt{Gemini-2.5-Flash} & 95.56 & 12.78 & 90.23 & 7.27 \\ 
\texttt{Gemini-2.5-Flash-Lite} & 95.56 & 8.89 & 90.00 & 1.47 \\ 
\texttt{DeepSeek-Chat} & 95.56 & 100.00 & 90.23 & 2.41 \\
\bottomrule
\end{tabular}
\caption{LLM backbone comparison on US filings.}
\label{tab:us_llm_cost}
\end{table}

\paragraph{Deployment of Backbone LLMs.}
Table~\ref{tab:us_llm_cost} compares different LLMs in the US Jurisdiction setting.
We can see that the overall fill rate is consistent across models, which suggests that coverage is driven primarily by the task pipeline rather than the choice of a backbone.
From a practical deployment perspective, the results indicate that smaller/efficient backbones can match the strongest models in terms of accuracy while being much cheaper, making them attractive default choices for real-world environments.

\section{The Demo Application}

\paragraph{Overview}

We develop a lightweight web-based demo for interactive exploration of financial statements extracted by \model. It targets financial analysts, cross-market investors, and financial NLP researchers requiring structured, auditable cross-jurisdiction reporting. Users select a market (US, JP, CN) and company, and the interface renders the three canonical statements—Income Statement (IS), Balance Sheet (BS), and Cash Flow (CF)—in tabbed views. Each view displays metadata such as fiscal year, filing date, currency, and accounting standard, and highlights key indicators including revenue, net income, total assets, and net operating cash flow.

\paragraph{Unified Schema and QA}

The outputs are normalized under a unified schema across markets while preserving market-specific labels and units. This enables consistent comparison without discarding local reporting semantics.
To support common analyst queries, the demo provides template-based question answering for high-frequency metrics, including revenue, net income, and operating cash flow. The QA module retrieves values from the normalized schema, enabling rapid validation without scanning raw worksheets.

\paragraph{Quality and Audit Signals}

Auditability is a primary design goal. We define a unified status ontology \texttt{OK}, \texttt{MISSING}, \texttt{PARSE\_ERROR}, or \texttt{NOT\_APPLICABLE} where \texttt{NOT\_APPLICABLE} is semantic rather than tied to literal wording in filings. The US/JP adapters predominantly instantiate OK/MISSING, while CN additionally activates \texttt{PARSE\_ERROR} and \texttt{NOT\_APPLICABLE} via PDF-oriented status tracing. Summary indicators report missing-field counts for quick completeness assessment. The interface also supports downloading structured workbooks for offline verification.

\paragraph{Implementation}

The demo is a no-database web service that reads structured Excel artifacts generated by the batch pipeline, ensuring consistency with experimental results. The backend is implemented in Python, and the frontend is a lightweight static page with API endpoints for market listing, company selection, statement retrieval, QA, and workbook download.
The demo can be launched locally with a single command and accessed remotely via SSH port forwarding, enabling reproducible presentations.

\paragraph{Demo Flow}

A typical session proceeds as follows: (1) select market and company, (2) inspect IS/BS/CF tabs and metadata, (3) issue QA prompts for core metrics, (4) download the workbook for manual cross-checking. This workflow emphasizes usability, explicit quality signals, and direct support for human verification.

\section{Conclusion and Future Work}

We presented \model, an agentic workflow for cross-jurisdiction financial reporting. By canonicalizing heterogeneous filings into a unified IS/BS/CF representation and decomposing processing into auditable stages with explicit verification signals, \model addresses key frictions in global financial analysis. Unlike free-form LLM extraction pipelines, \model uses the LLM as a constrained verifier, repairing outputs only when sufficient evidence is available and deferring uncertain cases to human review. This design enables transparent cross-market reporting, surfaces aggregation and mapping anomalies, and provides a scalable, auditable infrastructure that reduces reliance on manual data curation and mitigates hidden structural errors downstream.

Future work will expand jurisdiction coverage and enrich the canonical ontology to capture long-tail taxonomy variations, extend verification to footnotes and segment disclosures, and improve robustness across statements, periods, and noisy PDFs.

\section*{Limitations}

We acknowledge limitations of \model.
First, the current implementation focuses on annual filings from three jurisdictions (US, JP, and CN) and a fixed set of core IS/BS/CF targets. Extending to additional markets, reporting standards, and finer-grained line items remains future work.

Second, although XBRL-native jurisdictions provide structured inputs, PDF-centric environments (e.g., CN filings) introduce layout variability, fragmented tables, OCR noise, and issuer-specific conventions that reduce extraction reliability. In such cases, the system may defer to \textsc{Need\_Review}.

Third, canonical alignment relies on a predefined financial ontology, which may not fully capture market-specific nuances, long-tail taxonomy variations, or firm-specific disclosure idiosyncrasies.

Finally, \model\ is an auditable reporting assistant, not a fully autonomous system; human verification is necessary for high-stakes use cases.

\section*{Ethical Considerations}

\model operates in a high-stakes financial domain where mislocalized or incorrectly extracted values could mislead downstream analysis. To mitigate such risks, the system explicitly surfaces uncertainty through status labels, anomaly logs, and review flags, and constrains the LLM to bounded verification actions rather than free-form generation.

The demo system is intended for structured reporting assistance and inspection. It should not be used as the sole basis for investment or regulatory decisions without independent validation against original filings.

We also note that cross-jurisdiction localization necessarily involves modeling choices through a canonical ontology. While designed to preserve semantic faithfulness, any abstraction layer may introduce interpretative bias, especially in edge cases involving jurisdiction-specific accounting conventions. Users should therefore treat localized outputs as aligned representations rather than authoritative accounting restatements. Additionally, users are encouraged to review provenance metadata and cross-check anomalies to ensure transparency, traceability, and contextual correctness in interpretations.

\paragraph{Data License}

We use publicly available regulatory disclosures, including US filings from SEC EDGAR, Japanese filings from EDINET, and Chinese annual report PDFs from official public disclosure platforms. These documents are subject to the terms and conditions of their respective providers. 

Our released artifacts (e.g., code, canonical schemas, evaluation templates, structured outputs, and audit logs) do not include raw proprietary documents and can be redistributed independently of the original filings. Users are responsible for obtaining source filings directly from official regulatory portals in compliance with applicable data usage policies.
\bibliography{refs}

\clearpage
\newpage

\appendix



\end{document}